\documentclass[draftclsnofoot,onecolumn, 12pt]{IEEEtran}
\usepackage{etex}
\newcommand*{\QED}{%
\leavevmode\unskip\penalty9999 \hbox{}\nobreak\hfill
    \quad\hbox{$\square$}%
}
\newcommand*{\QEG}{%
\leavevmode\unskip\penalty9999 \hbox{}\nobreak\hfill
    \quad\hbox{$\clubsuit$}%
}
\newcommand*{\QDEF}{%
\leavevmode\unskip\penalty9999 \hbox{}\nobreak\hfill
    \quad\hbox{$\diamondsuit$}%
}

\usepackage[utf8]{inputenc} 
\usepackage{cite}
\usepackage{wrapfig} 
\usepackage[T1]{fontenc}
\usepackage{url}
\usepackage{ifthen}
\usepackage[cmex10]{amsmath} %

\usepackage{cleveref}
\usepackage[table]{xcolor}
\usepackage[cmex10]{amsmath}
\usepackage{amssymb}
\usepackage{amsthm}
\usepackage{mathtools}
\usepackage{booktabs}
\usepackage{bm}
\usepackage{graphicx}
\usepackage{url}
\IEEEoverridecommandlockouts
\usepackage{algorithm}
\usepackage{algorithmic}
\usepackage{balance}
\usepackage{multirow}
\usepackage{stmaryrd}
\newtheorem{theorem}{Theorem}
\newtheorem{remark}{Remark}
 
 \newtheorem{proposition}{Proposition}
 
 \newtheorem{lemma}{Lemma}
 
\theoremstyle{definition}

\newtheorem{assumption}{Assumption}

\theoremstyle{remark}

\newcommand{\diff}{\mathrm{d}}

\newcommand{\bI}{\mathbf{1}}

\usepackage{cite}

\DeclareMathOperator*{\argmin}{arg\,min}

\newcommand{\cP}{\mathcal{P}}

\newcommand{\bR}{\mathbb{R}}

\newcommand{\bQ}{\mathbb{Q}}

\newcommand{\bE}{\mathbb{E}}
\newcommand{\cW}{\mathcal{W}}
\newcommand{\Var}{\mathbb{V}\mathrm{ar}}
\newcommand{\Reg}{\varepsilon}
\newcommand{\orho}{\overline{\rho}}
\newcommand{\hP}{\widehat{\mathbb{P}}}
\newcommand{\bP}{\mathbb{P}}

\DeclarePairedDelimiterX{\inp}[2]{\langle}{\rangle}{#1, #2}
 
\newcommand{\trans}{^{\mathrm T}}

\title{
Non-Convex Robust Hypothesis Testing
using Sinkhorn Uncertainty Sets}
\begin{document}

\author{%
  \IEEEauthorblockN{Jie~Wang,~Rui~Gao,~Yao~Xie}
 \thanks{
 J. Wang and Y. Xie are with H. Milton Stewart School of Industrial and Systems Engineering, Georgia Institute of Technology. R. Gao is with Department of Information, Risk, and Operations Management, University of Texas at Austin.
 }
  }

\maketitle

\begin{abstract}
We present a new framework to address the non-convex robust hypothesis testing problem, wherein the goal is to seek the optimal detector that minimizes the maximum of worst-case type-I and type-II risk functions.
The distributional uncertainty sets are constructed to center around the empirical distribution derived from samples based on Sinkhorn discrepancy.
Given that the objective involves non-convex, non-smooth probabilistic functions that are often intractable to optimize, existing methods resort to approximations rather than exact solutions.
To tackle the challenge, we introduce an exact mixed-integer exponential conic reformulation of the problem, which can be solved into a global optimum with a moderate amount of input data.
Subsequently, we propose a convex approximation, demonstrating its superiority over current state-of-the-art methodologies in literature.
Furthermore, we establish connections between robust hypothesis testing and regularized formulations of non-robust risk functions, offering insightful interpretations.
Our numerical study highlights the satisfactory testing performance and computational efficiency of the proposed framework. 
\end{abstract}

\section{Introduction}

Hypothesis testing is a fundamental problem in statistics, whose primary goal is to decide the true hypothesis while minimizing the risk of wrong decisions. Hypothesis testing is a building block for various statistical problems such as change-point detection~\cite{Vincent08,xie2020sequential,Liyanchange_21,xie2022minimaxdec,xie2023distributionally}, model criticism~\cite{lloyd2015statistical, chwialkowski2016kernel, binkowski2018demystifying}, and it has applications in broad domains including healthcare~\cite{schober2019two}.
In some real-world applications, the underlying true distributions corresponding to each hypothesis are unknown, and we only have access to a small amount of data collected for each hypothesis.

Distributionally robust hypothesis testing has emerged as a popular approach to tackle the challenge of establishing an optimal decision in the presence of limited sample size, model misspecification, and adversarial data perturbation.
It formulates the problem as seeking the optimal decision over uncertainty sets that contain
candidate distributions for each hypothesis.
The construction of such distributional uncertainty sets plays a key role in both computational tractability and testing performance.

The earlier work of finding a robust detector dates back to Huber's seminar work~\cite{Peter65}, which constructs the uncertainty sets as total-variation probability balls centered around the reference distributions.
Unfortunately, the computational complexity of seeking the optimal detector within this framework, particularly for multivariate distributions, hinders its practical applications. Two primary approaches have emerged for constructing uncertainty sets in robust hypothesis testing. 
The first involves defining uncertainty sets using descriptive statistics such as moment conditions~\cite{magesh2023robust}.
The second approach considers all possible distributions within a pre-specified statistical divergence from a reference distribution. Commonly adopted statistical divergences include the KL-divergence~\cite{Levy09, gul2017minimax}, Wasserstein distance~\cite{gao18robust, xie2021robust, xie2022minimax}, entropic regularized 
Wasserstein distance~(i.e., Sinkhorn discrepancy)~\cite{wang2022data}, and maximum mean discrepancy~\cite{sun2023kernel}.

It is noteworthy that distributionally robust optimization (DRO) with Sinkhorn discrepancy-based uncertainty set has recently received great attention in the literature~\cite{wang2021sinkhorn, azizian2023regularization, wang2022improving, yang2023distributionally, wang2022data, dapogny2023entropy, song2023provably}, mainly due to its data-driven nature, computational tractability, and flexibility to obtain worst-case distributions yielding satisfactory performance.
Considering its empirical success, we propose a new framework for robust hypothesis testing, whose distributional uncertainty sets are constructed using the Sinkhorn discrepancy. 
Our goal is to seek the optimal detector to minimize the \emph{maximum of worst-case type-I and type-II error}.
In contrast to the recent works~\cite{wang2022data, gao18robust}  considering a special \emph{smooth} and \emph{convex} relaxation of the objective function, we aim to solve the non-convex problem by (i) either \emph{directly} optimizing the probabilistic objective, or (ii) providing a \emph{tighter} convex relaxation.

Our proposed framework balances the trade-off between computational efficiency and statistical testing performance. The contributions are summarized as follows.
\begin{enumerate}
    \item
Under the random feature model, we obtain a finite-dimensional optimization reformulation for this robust hypothesis testing problem~(Section~\ref{Section:setup}).
Besides, we provide a closed-form expression for the worst-case distributions~(Remark~\ref{remark:worst}).
    \item
We develop novel optimization algorithms for the non-convex robust testing problem. First, we provide an exact mixed-integer conic reformulation of the problem~(Section~\ref{Sec:MIP}), enabling the attainment of global optimum even with a moderate data size. Subsequently, we introduce a convex approximation and illustrate its superiority as a tighter relaxation compared to the state-of-the-art~(Section~\ref{Sec:cvx}).
\item 
We connect robust hypothesis testing and regularized formulations of non-robust risk functions under two hyper-parameter scaling regimes, offering insightful interpretations of Sinkhorn robust testing~(Section~\ref{Sec:regularization}).
\end{enumerate}
Our numerical study highlights the good statistical testing performance and the proposed framework's computational efficiency. Proofs and additional numerical study details can be found in the Appendix.\\
\noindent{\it Notations.}
The base of the logarithm function $\log$ is $e$. 
For any positive integer $N$, define $[N]=\{1,\ldots,N\}$.
For scalar $x\in\mathbb{R}$, define $(x)_+=\max\{x,0\}$.
Let $\mathcal{K}_{\exp}$ denote the exponential cone: 
\[
\mathcal{K}_{\exp}=\Big\{
(\nu,\mu,\delta)\in \bR_+\times\bR_+\times\bR:~
e^{\delta/\nu}\le \mu/\nu
\Big\}.
\]
For a given event $E$, define the indicator function $\mathbf{1}_E(\cdot)$ such that $\mathbf{1}_E(z)=1$ if $z\in E$ and otherwise $\mathbf{1}_E(z)=0$.
Given a function $\phi:~\Omega\to\mathbb{R}$ and scalar $r\in[1,\infty]$, define the norm $\|\phi\|_{L^r}=\left(\int_{\Omega}\phi(x)^r \diff x\right)^{1/r}$.
For a non-negative measure $\nu$, define the norm $\|\phi\|_{L^r(\nu)}=\left(\int_{\Omega}\phi(x)^r\diff\nu(x)\right)^{1/r}$.

\section{Problem Setup}\label{Section:setup}

Let $\Omega\subseteq\mathbb{R}^d$ be the sample space where the observed samples take their values, and $\cP(\Omega)$ be the set of all distributions supported on $\Omega$.
Denote by $\cP_1, \cP_2\subseteq\cP(\Omega)$ the uncertainty sets under hypotheses $H_1$ and $H_2$, respectively. 
Given two sets of training samples $\{x_1^k,\ldots,x_{n_k}^k\}$ generated from $\bP_k\in\cP_k$ for $k=1,2$, denote the corresponding empirical distributions as 
\[\hP_k=\frac{1}{n_k}\sum_{i=1}^{n_k}\delta_{x_i^k}.\]
For notation simplicity, assume that $n_0=n_1=n$, but our formulation can be naturally extended for unequal sample sizes.
Given a new testing sample $\omega$, the goal of \emph{composite hypothesis testing} is to distinguish between the null hypothesis $H_1:~\omega\sim \bP_1$ and the alternative hypothesis $H_2:~\omega\sim \bP_2$, where $\bP_k\in\cP_k$ for $k=1,2$.
For a detector $T:~\Omega\to\bR$, it accepts the null hypothesis $H_1$ when $T(\omega)\ge0$; otherwise, it accepts the alternative hypothesis $H_2$.
Under the Bayesian setting, we quantify the risk of this detector as the \textit{maximum of the worst-case type-I and type-II errors}:
\[
 \mathcal{R}(T;\cP_1,\cP_2)=\max\Big(\sup_{\bP_1\sim \cP_1}~\bP_1\{\omega: T(\omega)<0\},\sup_{\bP_2\sim \cP_2}~\bP_2\{\omega: T(\omega)\ge0\}\Big).
\]
In this paper, we aim to find the detector $T$ such that its risk is minimized:
\begin{equation}
\label{Eq:testing}
\inf_{T:~\Omega\to\mathbb{R}}~
\mathcal{R}(T;\cP_1,\cP_2).
\end{equation}
It is worth noting that there are three major challenges when solving such a formulation:
(i) First, seeking the optimal detector among all measurable functions is an infinite dimensional optimization problem, which is intractable;
(ii) Second, finding the worst-case distributions over ambiguity sets $\cP_1, \cP_2$ is also an infinite dimensional optimization, which is not always tractable;
(iii) Finally, the objective involves probability functions, which are non-smooth and non-convex.
In the following, we provide methodologies to tackle these difficulties.

\subsection{Random Feature Model}

We use the random feature model to address challenges ~(i). 
We make the following assumption regarding the space of detectors.
\begin{assumption}[Optimal Detector in RKHS]\label{Assumption:detector}
The underlying true detector $T^*:~\Omega\to \mathbb{R}$ belongs to a reproducing kernel Hilbert space~(RKHS) $\mathcal{F}_K$ equipped with a kernel function $K(x,y) = \mathbb{E}_{\omega\sim \pi_0}[\phi(x; \omega)\phi(y; \omega)]$ for some feature map $\phi$ and feature distribution $\pi_0$.
Besides, there exists a constant $M>0$ such that for $\pi_0$-almost $\omega$, it holds that $\|\phi(\cdot;\omega)\|_{L^2}\le M$.\QEG
\end{assumption}
We highlight that such a restriction does not limit the generality.
Instead, since the RKHS (with universal kernel choice, such as Gaussian kernel) is dense in the continuous function space, i.e., it approximates any continuous function within arbitrarily small error.
For commonly used kernels, the feature map expressions are also easily satisfied.
For example, when considering the kernel function to be continuous, real-valued, and shift-invariant, by Bochner’s
Theorem~\cite{rudin2017fourier}, it holds that 
\[
K(x,y) = \mathbb{E}_{(z,b)}
[\cos(z\trans x+b)\cos(z\trans y+b)],
\]
where the vector $z$ follows the distribution from the density function $p(\omega) = \frac{1}{2\pi}\int e^{-\mathbf{i}\inp{\omega}{\delta}}K(\delta)\diff\delta$,  $\mathbf{i} = \sqrt{-1}$, and scalar $b$ follows the uniform distribution supported on $[0,2\pi]$.
In such case, Assumption~\ref{Assumption:detector} holds by taking $\phi(x;\omega):=\cos(z\trans x+b)$ with $\omega:=(z,b)$.

For any detector $T\in\mathcal{F}_K$, there exists a function $\theta(\cdot)\in L^2(\pi_0)$ such that 
\[
T(x) = \mathbb{E}_{\omega\sim \pi_0}[\theta(\omega)\phi(x;\omega)].
\]
Denote the feature vector
\[
\Phi(x)=\left(\frac{1}{D}\phi(x;\omega_1),\ldots,\frac{1}{D}\phi(x;\omega_D)\right)\in\mathbb{R}^D,
\]
with $\{\omega_i\}_{i\in[D]}$ being i.i.d. samples generated from $\pi_0$, and the vector $\bar{\theta}=(\theta(\omega_1),\ldots,\theta(\omega_D))$.
Then, the random feature model 
\[
\hat{T}(x) = \inp{\bar{\theta}}{\Phi(x)}=\frac{1}{D}\sum_{i\in[D]}~\theta(\omega_i)\phi(x;\omega_i)
\]
is an unbiased estimator of $T(x)$ with respect to $\{\omega_i\}$.
This motivates us to propose 
\[
\mathcal{F}_D = \left\{
T:~x\mapsto \inp{\theta}{\Phi(x)},~
\exists
\theta\in\mathbb{R}^D
\right\}
\]
as an approximation of $\mathcal{F}_K$.
The following presents the approximation theoretical guarantees for $\mathcal{F}_D$.
Similar theoretical results have also been explored in recent literature~\cite[Theorem~5]{ma2020towards}.
\begin{proposition}[Direct Approximation Theorem]
\label{Pro:direct:app}
Fix the error probability $\delta\in(0,1)$.
Suppose Assumption~\ref{Assumption:detector} holds and define the norm
\[
\|T^*\|_{\infty}:=\inf_{\theta(\cdot)}\left\{ 
\|\theta(\cdot)\|_{L^{\infty}(\pi_0)}:~
T^*(x) = \mathbb{E}_{\pi_0}[\theta(\omega)\phi(x;\omega)]
\right\}.
\]
Then there exists a function $T$ in $\mathcal{F}_D$ such that with probability at least $1-\alpha$, it holds that
\begin{equation}
\|T^* - T\|_{L^2}\le 
\frac{M}{\sqrt{D}}
\left( 
\|T^*\|_{\mathcal{H}_K}
+
\sqrt{2\|T\|_{\infty}^2\log\frac{1}{\alpha}}
\right).
\tag*{\QDEF}
\end{equation}
\end{proposition}
In other words, the number of samples $D=\Omega(\frac{1}{\epsilon^2}\log\frac{1}{\alpha})$ is enough to control the approximation error within $\epsilon$ with probability at least $1-\alpha$, which is \emph{data dimension independent}.
This justifies that the random feature model is a suitable choice for approximate detectors, especially for high-dimensional scenarios.
As a consequence, we obtain the finite-dimensional reformulation of Problem~\eqref{Eq:testing}:
\begin{equation}
\label{Eq:CCP:testing}
\begin{aligned}
\min_{\theta\in\mathbb{R}^D, s\ge0}&\quad s\\
\mbox{s.t.}\quad 
\sup_{\bP_k\in\cP_k}~\bP_1\{&\omega:~(-1)^{k+1}\inp{\theta}{\Phi(\omega)}<0\}\le s,~~k=1,2.
\end{aligned}    
\end{equation}
From the formulation above, we realize that $\theta$ is optimal, implying $\alpha \theta$ is also optimal for any scalar $\alpha\in\mathbb{R}_+$.
To make the solution $\theta$ well-conditioned, we additionally add the constraint $\|\theta\|_2\le 1$ when solving \eqref{Eq:CCP:testing}.

\subsection{Preliminaries on Sinkhorn DRO}
In the following, we specify uncertainty sets $\mathcal{P}_k, k=1,2$ using Sinkhorn discrepancy and discuss the corresponding tractable reformulation.%
\begin{assumption}[Sinkhorn Uncertainty Sets]\label{Assumption:Sinkhorn}
For $k=1,2$, we specify the uncertainty set
\begin{equation}
\mathcal{P}_k = \Big\{ 
\bP:~\cW_{\Reg_k}(\bP, \hP_k)\le \rho_k
\Big\},
\label{Eq:bP:k}
\end{equation}
where the Sinkhorn discrepancy $\cW_{\Reg}(\cdot,\cdot)$ is defined as 
\[
\mathcal{W}_{\Reg}(\bP, \bQ) \:= \inf_{
\substack{
\gamma\in\Gamma(\bP, \bQ)
}}~ \left\{\mathbb{E}_{(x,y)\sim\gamma}[c(x,y)] + \Reg H(\gamma)\right\}.
\]
Here $\Gamma(\bP,\bQ)$ denotes the set of joint distributions whose first and second marginal distributions are $\bP$ and $\bQ$ respectively, $c(x,y)$ denotes the transport cost, and $H(\gamma)$ denotes the relative entropy of $\gamma$:%
\begin{equation}
\tag*{\QEG}
H(\gamma)\:= \bE_{(x,y)\sim\gamma}\left[\log\left( 
\frac{\diff\gamma(x,y)}{\diff\gamma(x)\diff y}
\right)\right].
\end{equation}
\end{assumption}
Subsequently from Assumption~\ref{Assumption:Sinkhorn}, we define the radii
\begin{equation}\label{Eq:radii}
\orho_k\triangleq \rho_k+\mathbb{E}_{x\sim\hP}\Big[\Reg_k\log
\int e^{-c(x,z)/\Reg_k}\diff z\Big], \quad k=1,2.
\end{equation}
With a measurable variable $f:~\Omega\to\bR$, we associate value
\begin{equation}\label{Eq:primal:Sinkhorn}
V = \sup_{\bP\in\cP}~\mathbb{E}_{\bP}[f],
\end{equation}
where the ambiguity set $\cP$ is in the form of \eqref{Eq:bP:k}.
Define the dual problem of \eqref{Eq:primal:Sinkhorn} as 
\begin{equation}\label{Eq:dual:Sinkhorn}
V_D = \inf_{\lambda\ge0}~\left\{\lambda\bar{\rho} + 
\mathbb{E}_{x\sim\hP}\Big[\lambda\Reg\log
\mathbb{E}_{z\sim\bQ_{x,\Reg}}\big[ 
e^{f(z)/(\lambda\Reg)}
\big]\Big]
\right\},
\end{equation}
where we define the constant 
\begin{equation}
\orho=\rho + \mathbb{E}_{x\sim\hP}\Big[\Reg\log\int e^{-c(x,z)/\Reg}\diff z\Big]\label{Eq:def:orho}
\end{equation}
and $\bQ_{x,\Reg}$ as the kernel probability distribution with density 
\begin{equation}\label{Eq:bQ}
\frac{\diff\bQ_{x,\Reg}(z)}{\diff z}\propto e^{-c(x,z)/\Reg}.
\end{equation}
For example, when considering the optimal transport cost function $c(x, z)=\frac{1}{2}\|x-z\|_2^2$, $\bQ_{x,\Reg}$ reduces to the Gaussian distribution $\mathcal{N}(x,\Reg \mathbf{I}_D)$.
By \cite[Theorem~1]{wang2021sinkhorn}, $V_{D}$ defined in \eqref{Eq:dual:Sinkhorn} is the dual reformulation of Problem~\eqref{Eq:primal:Sinkhorn}.
This observation indicates the computational tractability when using Sinkhorn uncertainty sets: solving the worst-case expectation problem in \eqref{Eq:primal:Sinkhorn} is \emph{always} tractable when solving its one-dimensional dual problem in \eqref{Eq:dual:Sinkhorn} using the random sampling approach developed in \cite[Section~4]{wang2021sinkhorn}.
Besides, we usually tune the radius $\orho$ that appeared in dual formulation instead of the original radius $\rho$.
\vspace{-0.5em}
\begin{proposition}[Reformulation of Sinkhorn DRO]\label{Theorem:Sinkhorn}
Suppose that $
\int e^{-c(x,z)/\Reg}\diff z<\infty$ for $\hP$-almost every $x$ and $\orho\ge0$, then it holds that $V = V_D$.\QDEF
\end{proposition}

\begin{remark}[Recovery of Worst-case Distributions]\label{remark:worst}
After solving Problem~\eqref{Eq:CCP:testing} to obtain (near-)optimal solution $(\theta^*, s^*)$, one can recover the worst-case distributions corresponding to hypothesis $H_1$ and $H_2$ based on \cite[Remark~4]{wang2021sinkhorn}, denoted as $\bP_k^*, k\in\{1,2\}$.
Assume the optimal Lagrangian multipliers $(\lambda_k^*)_{k=1,2}$ to Problem~\eqref{Eq:CCP:testing} is positive,
then the density of the worst-case distribution becomes
\[
\frac{\diff\bP_k(z)}{\diff z}
=
\bE_{x\sim\hP_k}\Big[ 
\alpha_x\cdot \exp\left(
\frac{\bI \{ (-1)^{k+1}\inp{\theta^*}{\Phi(z)}<0 \} - \lambda_k^*c(x,z)}{\lambda_k^*\Reg_k}
\right)
\Big],
\]
where $\alpha_x$ is a normalizing constant.
\QED
\end{remark}

\section{Optimization Methodology}
\label{Sec:methodology}

In this section, we first discuss how to solve the formulation \eqref{Eq:CCP:testing} directly based on mixed-integer conic programming and then talk about how to solve its convex relaxation using the CVaR approximation approach.

\subsection{A Mixed-Integer Conic Formulation}\label{Sec:MIP}
According to the definition of ambiguity sets $\cP_k, k=1,2$ and Proposition~\ref{Theorem:Sinkhorn}, 
as probabilistic constraints can always be written as expectations of indicator functions,  Problem~\eqref{Eq:CCP:testing} can be reformulated as 
\begin{equation}
\label{Eq:CCP:testing:duality}
\vartheta^*=
\min_{
\substack{
\|\theta\|_2\le 1, 
s\ge0, \\
\lambda_1, \lambda_2\ge0
}}~\Big\{ 
s:~F_k(\theta, \lambda_k)\le s, \quad k = 1, 2
\Big\},
\end{equation}
where the function $F_k, k\in\{1,2\}$ is defined as
\[
F_k(\theta, \lambda_k)\triangleq
\lambda_k\orho_k +\bE_{x\sim \hP_k}\left[ 
 \lambda_k\Reg_k\log\bE_{y\sim \bQ_{x,\Reg_k}}\left[ 
 \exp\left\{\frac{\bI \{(-1)^{k+1}\inp{\theta}{\Phi(y)}<0 \}}{\lambda_k\Reg_k}\right\}
\right]
\right].
\]
Here, the radii $\orho_k$ and distribution $\bQ_{x, \Reg_k}$ are defined in \eqref{Eq:radii} and \eqref{Eq:bQ}, respectively.
Next, we adopt the idea of sample average approximation~(SAA) to approximate those two constraints in \eqref{Eq:CCP:testing:duality}. 
Recall that $\hP_k=\frac{1}{n}\sum_{i=1}^n\delta_{x_i^k}, k\in\{1,2\}$.
For each sample $x_i^k$, we generate $m$ i.i.d. sample points $y_{i,j}^k$ following distribution $\bQ_{x_i^k}$ for $j\in[m]$.
Hence, we obtain the sample estimates of functions $F_k$ for $k=1,2$:
\[
\widetilde{F}_k(\theta, \lambda_k)=
\lambda_k\orho_k + \frac{\lambda_k\orho_k}{n}\sum_{i\in[n]}
\log\left[ \frac{1}{m}\sum_{j\in[m]}
e^{\frac{\bI \{(-1)^{k+1}\inp{\theta}{\Phi(y_{i,j}^k)}<0 \}}{\lambda_k\Reg_k}}
\right].
\]
Consequently, the sample estimate of the optimal value $\vartheta^*$ defined in \eqref{Eq:CCP:testing:duality} is given by
\begin{equation}
\widehat{\vartheta}^*=\min_{
\substack{
\|\theta\|_2\le 1, 
s\ge0, \\
\lambda_1, \lambda_2\ge0
}}~\Big\{ 
s:~\widetilde{F}_k(\theta, \lambda_k)\le s, \quad k=1,2
\Big\}.\label{Eq:CCP:testing:duality:app}
\end{equation}
We present a consistency result between $\widehat{\vartheta}^*$ and $\vartheta^*$ below.
\begin{proposition}[Consistency of $\widehat{\vartheta}^*$]\label{Pro:consistent}
Assume the radii $\orho_k>0$ for $k=1,2$,
and there exists an optimal solution $(\theta^*, s^*, \lambda_1^*, \lambda_2^*)$ to \eqref{Eq:CCP:testing:duality} such that for any $\delta>0$, there exists $(\theta, s, \lambda_1, \lambda_2)$ with $\|\theta\|_2\le 1, s\ge0, \lambda_1\ge0, \lambda_2\ge0$, $\|(\theta, s, \lambda_1, \lambda_2) - (\theta^*, s^*, \lambda_1^*, \lambda_2^*)\|\le \delta$ and $F_k(\theta, \lambda_k)<s^*, k=1,2$.
As a consequence, $\widehat{\vartheta}^*\to \vartheta^*$.
\QDEF
\end{proposition}

The assumptions in Proposition~\ref{Pro:consistent} are essential.
The first condition is to ensure the optimal multipliers $\lambda_1,\lambda_2$ exist and are bounded.
For the second condition, assume on the contrary that there exists a case where $F_k(\theta, \lambda_k)\le s^*$ only defines one feasible point $(\bar{\theta}, \bar{\lambda}_k)$ such that $F_k(\bar{\theta}, \bar{\lambda}_k)=s^*$.
Then arbitrarily small perturbations regarding the constraint $\widetilde{F}_k(\theta, \lambda_k)\le s^*$ may cause the SAA problem~\eqref{Eq:CCP:testing:duality:app} becoming infeasible to solve.

Besides, the SAA problem~\eqref{Eq:CCP:testing:duality:app} admits a finite-dimensional mixed-integer exponential conic program~(MIECP) reformulation.
Consequently, the moderate-sized instances of such a formulation could be handled by state-of-the-art solvers~\cite{dahl2022primal, coey2020outer, ye2021second} in a reasonable amount of time.

\begin{theorem}[MIECP Reformulation of \eqref{Eq:CCP:testing:duality:app}]\label{Thm:MIECP}
Assume there exist constants for $i\in[n],j\in[m],k\in\{1,2\}$:
\[
M_{i,j}^k=\max_{\|\theta\|_2\le 1}~(-1)^{k+1}\inp{\theta}{\Phi(y_{i,j}^k)}.
\]
Then, Problem~\eqref{Eq:CCP:testing:duality:app} is equivalent to %
\begin{equation}\label{Eq:MIECP}
\begin{aligned}
\mbox{\rm Minimize}&\quad s\\
\mbox{s.t.}&\quad \left\{
\begin{aligned}
\|\theta\|_2&\le 1\\
(-1)^{k+1}\inp{\theta}{\Phi(y_{i,j}^k)}&\le M_{i,j}^k(1-z_{i,j}^k)%
\end{aligned}
\right.
\\
&\quad 
\left\{
\begin{aligned}
\lambda_k\orho_k + \frac{1}{n}\sum_{i\in[n]}t_i^k&\le s\\
\lambda_k\Reg_k&\ge \frac{1}{m}\sum_{j\in[m]}a_{i,j}^k\\
(\lambda_k\Reg_k, a_{i,j}^k, z_{i,j}^k - t_{i}^k)&\in\mathcal{K}_{\exp},\\
&\quad  i\in[n],j\in[m],k\in\{1,2\}
\end{aligned}
\right.\\
\end{aligned}
\end{equation}
subject to the following decision variables
\begin{equation}
\tag*{\QDEF}
\begin{multlined}
s\in[0,1],\theta\in\bR^D, \lambda_1,\lambda_2\in \mathbb{R}_+, \{t_i^k\}_{i,k}\in\mathbb{R}^{n\times 2},\\
\{z_{i,j}^k\}_{i,j,k}\in\{0,1\}^{n\times m\times 2},
\{a_{i,j}^k\}_{i,j,k}\in\mathbb{R}^{n\times m\times 2}.
\end{multlined}
\end{equation}
\end{theorem}

Although Problem~\eqref{Eq:MIECP} can be directly handled by off-the-shelf Mosek solver~\cite{aps2019mosek}, we do not implement in this way because it involves $2nm$ binary variables and $2nm$ exponential conic constraints, which incurs heavy computational cost.
Instead, we solve it using the outer approximation algorithm developed in \cite{coey2020outer}, which iteratively solves the subproblem of \eqref{Eq:MIECP} for fixed values of binary variables $\{z_{i,j}^k\}_{i,j,k}$ and then update them using the cutting plane algorithm.

\subsection{Convex Approximation}\label{Sec:cvx}
Since the probabilistic constraints in Problem~\eqref{Eq:CCP:testing} make it intractable to solve, an alternative approach to solving this problem is to construct convex approximations of those constraints.
The most popular approach is to replace the probabilistic constraints with the conditional value-at-risk~(CVaR) approximation~\cite{nemirovski2007convex}, since the following relation holds for any random variable $Z$ and probability level $\epsilon$: 
\[
\inf_{\beta\le0}~\Big\{ 
\epsilon\beta + \mathbb{E}[Z-\beta]_+
\Big\}\le 0\implies 
\bP\{Z>0\}\le \epsilon.
\]
Inspired by this approach, we replace two constraints in Problem~\eqref{Eq:CCP:testing} using the CVaR approximation:
\begin{subequations}
\begin{align}
\min_{\|\theta\|_2\le 1, s\ge0}&\quad s\\
\mbox{s.t.}\quad
\sup_{\bP_k\in\cP_k}\inf_{\beta_k\le0}\Big\{&
s\beta + \mathbb{E}_{\bP_k}[(-1)^k\inp{\theta}{\Phi(\omega)} - \beta_k]_+
\Big\}\le0,\quad k=1,2.\label{Eq:CVaR:b}
\end{align}
\end{subequations}
\begin{remark}[Superior Performance of CVaR Approximation]\label{Remark:superior}
Recall references~\cite{gao18robust, wang2022data} used the generating function approach for convex approximation, which can be viewed as a special case of our formulation by specifying $\beta_k=-1, \forall k$ in \eqref{Eq:CVaR:b}.
In such cases, this constraint becomes
\[
\sup_{\bP_k\in\cP_k}~\mathbb{E}_{\bP_k}\left[
\ell\circ\Big((-1)^k\inp{\theta}{\Phi(\omega)}\Big)
\right]\le s,\quad k=1,2,
\]
with the generating function $\ell(x)=(x+1)_+$ that leads to the tightest theoretical approximation ratio proposed in \cite[Theorem~1]{gao18robust}.
Their approaches can be strengthened by taking the optimization over $\beta_k$ into account.\QED
\end{remark}
Assume the sample space $\Omega$ is compact. By Prohorov’s Theorem~(see, e.g., \cite[Theorem~2.4]{van2000asymptotic}), the ambiguity sets $\cP_k, k=1,2$ are compact as well, which ensures that one can apply Sion’s minimax Theorem~\cite{sion1958general} to exchange the sup and inf operators in those two constraints of the problem above.
Next, one can leverage the strong duality result in Proposition~\ref{Theorem:Sinkhorn} to obtain its equivalent formulation:
\begin{equation}\label{Eq:convex:relaxation}
\min_{\substack{\|\theta\|_2\le 1, s\ge0, \\
\beta_k\le 0, \lambda_k\ge0, k = 1, 2
}}~\Big\{
s:~G_k(s, \beta_k, \lambda_k)\le 0,\quad k=1,2
\Big\},
\end{equation}
where the function $G_k$ is defined as
\[
G_k(s,\beta_k,\lambda_k)=
s\beta_k + %
\Big\{ 
\lambda_k\orho_k + %
\bE_{x\sim \hP_k}\Big[ 
\lambda_k\Reg_k\log\bE_{y\sim \bQ_{x,\Reg_k}}\left[ 
e^{[(-1)^{k}\inp{\theta}{\Phi(y)} - \beta_k]_+/(\lambda_k\Reg_k)}
\right]
\Big\}.
\]
Here, the radii $\orho_k$ and distributions $\bQ_{x, \Reg_k}$ are defined in \eqref{Eq:radii} and \eqref{Eq:bQ}, respectively.
It is worth noting that Problem~\eqref{Eq:convex:relaxation} does not preserve convexity due to the bilinear structure of $(s,\beta_k)$ for $k=1,2$ in two constraints.
Fortunately, we can apply the bisection search method outlined in Algorithm~\ref{Alg:Sinkhorn} that finds the global optimum solution efficiently.
\begin{algorithm}[H]
\caption{
Bisection Search for Solving Problem~\eqref{Eq:convex:relaxation} 
} 
\label{Alg:Sinkhorn}
\begin{algorithmic}[1]\label{Alg:permutation:test}
\REQUIRE{Interval $[s^{\text{lb}}, s^{\text{ub}}]$, precision level $\Upsilon$.}
\WHILE{$s^{\text{ub}} - s^{\text{lb}} < 
\Upsilon$} 
\STATE{$s\leftarrow \frac{1}{2}(s^{\text{lb}}+s^{\text{ub}})$.}
\STATE{Compute
\vspace{-1em}
\begin{equation}
\label{Eq:opt:feasibility:sub}
T(s)=\min_{\substack{\theta\in\mathbb{R}^D, \|\theta\|_2\le 1, \beta_k\le 0, \\
\lambda_k\ge0, k=1,2
}}\left\{
\max_{k}~G_k(s,\beta_k,\lambda_k)
\right\}.
\end{equation}
}
\STATE{Update $s^{\text{ub}}\leftarrow s$ \textbf{if} $T(s)\le 0$ and \textbf{otherwise} $s^{\text{lb}}\leftarrow s$.}
\ENDWHILE\\
\textbf{Return} $s$
\end{algorithmic}
\end{algorithm}
The most computationally expansive step in Algorithm~\ref{Alg:Sinkhorn} is to solve the subproblem ~\eqref{Eq:opt:feasibility:sub}.
One can apply the projected stochastic subgradient method~\cite{nemirovski2009robust} to obtain its optimal solution with a negligible optimality gap.
The main difficulty is obtaining unbiased gradient estimates of $G_k$ since the objective function involves nonlinear operators of expectations.
Instead, one can follow the approach outlined in \cite{hu2021bias, hu2020biased, hu2023contextual} to efficiently generate biased gradient estimates with controlled gradient bias and variance.
Consequently, one can still obtain the optimal solution with convergence guarantees. 
We leave the complexity analysis of this method for future study.

It is also noteworthy that CVaR approximation has been used to solve the Sinkhorn robust chance-constrained program in literature~\cite{yang2023distributionally}.
Unlike their algorithm idea that solves a large-scale convex program using interior-point methods, we provide a first-order method that enables us to solve such problem more efficiently.

\section{Regularization Effects of Robust Testing}\label{Sec:regularization}

Recall that we have used Sinkhorn ambiguity sets to robustify the probabilistic constraints in \eqref{Eq:testing}.
In this section, we provide interpretations of such kind of robustness by showing that the robust risk of a detector can be well approximated by the non-robust risk with certain regularizations, called the \emph{regularization effects}.

To begin with, we study the worst-case $0$-$1$ loss function for a generic event $E$ and a general nominal distribution $\hP$:
\begin{equation}
\sup_{\mathbb{P}:~\cW_{\Reg}(\bP, \hP)\le \rho}~\bP(E).\label{Eq:general:cW:E}
\end{equation}
We assume hyper-parameters $\orho$ defined in \eqref{Eq:def:orho} and regularization parameter $\Reg$ both converges to $0$, and we consider two scaling regimes between $\orho$ and $\eta$: either $\orho/\Reg\to0$ or $\Reg/\orho\to0$.

\noindent{\bf Case~1:~$\orho/\Reg\to0$.}
In this case, the convergence rate of the radius $\orho$ is faster than that of the regularization parameter $\Reg$.
Define the variance regularizer
$
\sigma^2(E;\hP,\Reg) = \mathbb{E}_{x\sim\hP}\Big[\Var_{z\sim\bQ_{x,\Reg}}[\mathbf{1}_E(z)]\Big].
$
The following proposition reveals that Problem~\eqref{Eq:general:cW:E} is asymptotically equivalent to the variance regularized $0$-$1$ loss, whose proof follows a similar argument from \cite{blanchet2023statistical}.
\begin{proposition}\label{Proposition:sinkhorn:reg}
For any $b_0>0$, the following holds for all $\Reg>0$ and Borel probability measures $\hP$ satisfying $\inf_{\Reg>0}\sigma^2(E;\hP,\Reg)\ge b_0$:
\begin{equation}
\begin{multlined}
\sup_{\mathbb{P}:~\cW_{\Reg}(\bP, \hP)\le \rho}~\bP(E)
-
\Big( 
\mathbb{E}_{x\sim\hP}[\bQ_{x,\Reg}(E)]
+
(2\orho/\Reg)^{1/2}\sigma(E; \hP, \Reg)
\Big)
=o((\orho/\Reg)^{1/2}).
\end{multlined}
\tag*{\QDEF}
\end{equation}
\end{proposition}
Based on Proposition~\ref{Proposition:sinkhorn:reg}, the objective in Problem~\eqref{Eq:testing} can be viewed as the variance-regularized non-robust testing problem with residual error $O(\max_{k=1,2}\orho_k/\Reg_k)$:
\[
\max_{k=1,2}\Bigg(
\hP_k(E_k)
+
(2\orho_k/\Reg_k)^{1/2}\sigma(E_k;\hP_k,\Reg_k)
\Bigg),
\]
where $E_1=\{\omega:~T(\omega)<0\}, E_2 = E_1^c$.

\noindent{\bf Case~2:~$\Reg/\orho\to0$.}
Next, we consider the case where the convergence rate of the entropic regularization $\Reg$ is faster than that of $\orho$.
To simplify the analysis, we consider the quadratic transport cost function $c(x,z)=\frac{1}{2}\|x-z\|_2^2$ for Sinkhorn discrepancy defined in Assumption~\ref{Assumption:Sinkhorn}.
In such a case, we show Problem~\ref{Eq:general:cW:E} is well approximated by the Wasserstein robust loss, whose proof is mainly based on Laplace's method~\cite{erdelyi1956asymptotic}.
\begin{proposition}\label{Pro:cW:W}
For any measurable subset $E\subseteq \Omega$, %
\begin{equation}
\sup_{\mathbb{P}:~\cW_{\Reg}(\bP, \hP)\le \rho}~\bP(E)
=
\sup_{\mathbb{P}:~\cW_{0}(\bP, \hP)\le \orho}~\bP(E) + O(\Reg/\orho),
\end{equation}
where $\cW_{0}(\cdot,\cdot)$ denotes the standard optimal transport distance with quadratic transport cost function.
\QDEF
\end{proposition}
We additionally assume that $\hP$ is an empirical distribution constructed from $n$ i.i.d. samples from the underlying true distribution $\bP_*$, and specify the radius $\orho=O(n^{-b})$ for some $b\in(0,1]$.
Based on Proposition~\ref{Pro:cW:W} and a recent study~\cite{yangwasserstein} that provides the regularization effect analysis on Wasserstein DRO with $0$-$1$ loss, we further expand Problem~\eqref{Eq:general:cW:E} as 
\[
\sup_{\mathbb{P}:~\cW_{\Reg}(\bP, \hP)\le \rho}~\bP(E)
=
\hP(E)+O(1)\cdot\mathfrak{g}(0)^{2/3}\orho^{2/3} + O(\Reg/\orho),%
\]
where the density $\mathfrak{g}(0):=\lim_{s\downarrow0}\frac{1}{s}\bP_*\{\omega: \mathsf{d}_{E^c}(\omega)\in(0,s)\}$.

Based on the argument above, the objective in Problem~\eqref{Eq:testing} can be viewed as the following density-regularized non-robust testing problem with residual error $O(\Reg/\orho)$:%
\[
\max_{k=1,2}\Bigg(
\hP_k(E_k)
+
O(1)\cdot \mathfrak{g}_k(0)^{2/3}\orho_k^{2/3}
\Bigg),
\]
where events $E_k=\{\omega:~(-1)^{k+1}\inp{\theta}{\Phi(\omega)}<0\}$, and density
\[
\mathfrak{g}_k(0)=\lim_{s\downarrow0}\frac{1}{s}\bP_k\{\omega\in E_k: 
\mathsf{d}_{E_k^c}(\omega)\in(0,s)\},\quad k=1,2.
\]
A large value of $\mathfrak{g}_k(0)$ means the detector has a small empirical margin around the decision boundary. Hence, the robust testing in the regime $\Reg/\orho\to0$ tends to penalize this density to penalize detectors with a small margin.

\section{Numerical Experiments}
In this section, we evaluate the performance of our robust hypothesis testing method based on exact MIECP reformulation or approximation algorithm and three other existing approaches.
The detailed implementation step for baseline approaches is omitted in Appendix~\ref{Appendix:B-A}.
In the following, we present a set of numerical examples using synthetic and real datasets.

Notably, the detector of Sinkhorn robust hypothesis testing is constructed using the random feature model.
In experiments, we construct our feature mapping $\Phi(\cdot)$ using the neural tangent kernel~(NTK) with the two-layer neural network with softplus activation, which has been studied in recent literature~\cite{cheng2021neural}.
Its benefits include light computational cost and satisfactory numerical performance. 
One exception is that we construct the NTK using a two-layer convolutional neural network for image datasets.
We take the number of entries in the feature mapping $\Phi(\cdot)$ as $D=100$.
Hyper-parameters that correspond to all approaches are tuned by cross-validation.
Specifically, we randomly partition the given samples into $50\%$-$50\%$ training and validation sets.
We compute detectors using training sets across different hyper-parameter choices and choose the one with the best performance using the validation set.
For SDRO model, we take regularization parameter $\Reg$ and radius $\orho$ from the set
\[
\Big\{
(\Reg,\orho):~\Reg\in\{\text{1e-2, 5e-2, 1e-1}\},\quad 
\orho\in \{\text{1e-1,5e-1,1}\}
\Big\}.
\]

\begin{figure}[ht]
    \centering
    \includegraphics[width=0.46\textwidth]{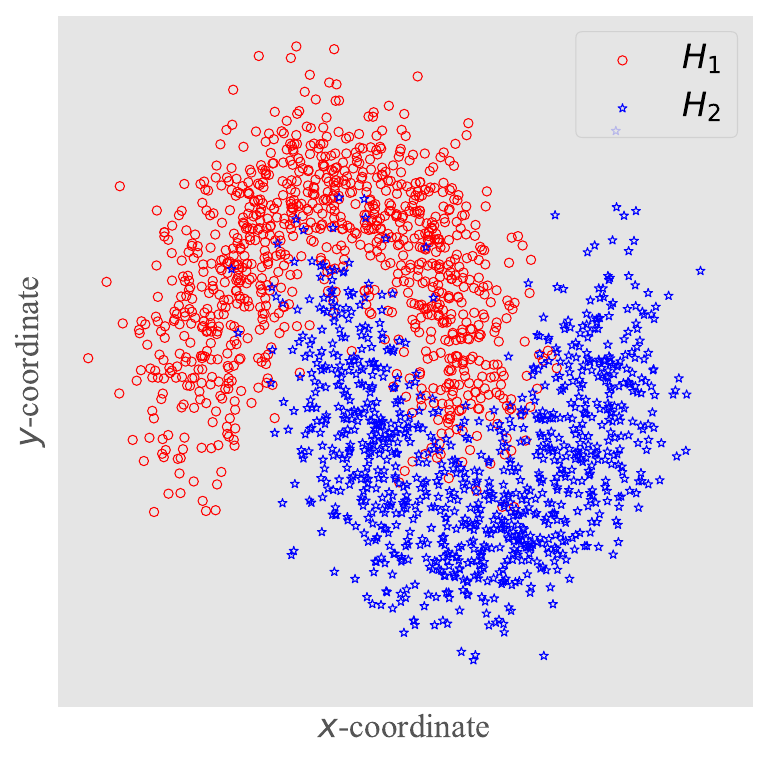}
    \includegraphics[width=0.46\textwidth]{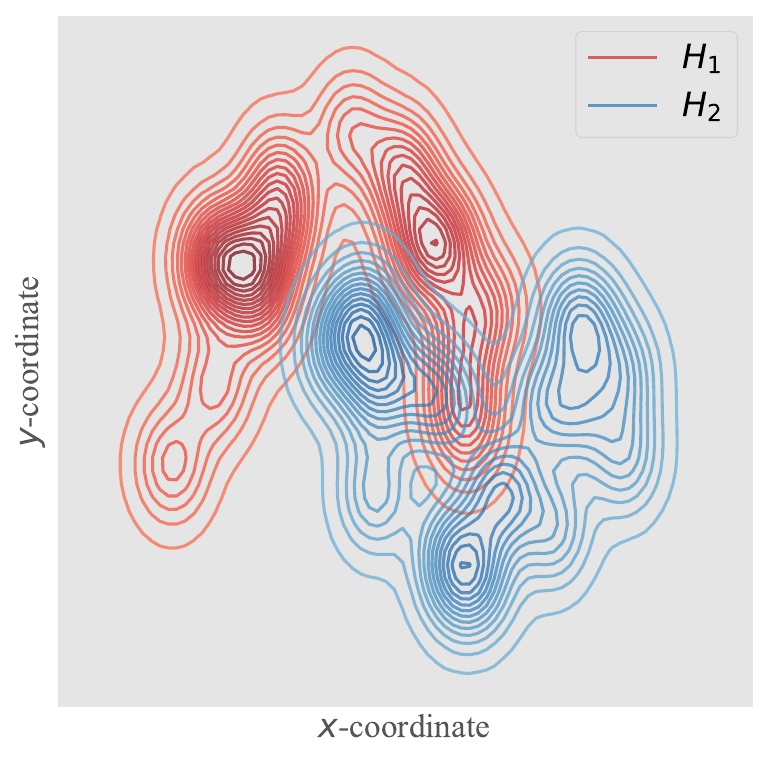}
    \caption{Visualization of worst-case distributions for $H_1$ and $H_2$, respectively, on noisy Moon Dataset. Left: scatter plot for data points with testing size $n_{\text{Te}}=1000$; right: visualization of worst-case distributions based on training samples. The maximum of type-I/type-II error is $\mathsf{0.165}$, with computational time $264.1$s.}
    \label{fig:moon}
\end{figure}

We first provide visualization of the worst-case distributions for Sinkhorn robust hypothesis testing using its MIECP reformulation with noisy Moon dataset~\cite{scikit-learn} and training size $n=20$ in Fig.~\ref{fig:moon}.
From the plot, we can see our Sinkhorn testing framework presents a data-driven way to obtain the worst-case distributions that reasonably match the ground truth.

\begin{figure}[ht]
\vspace{1em}
    \centering
   \includegraphics[width=0.8\textwidth]{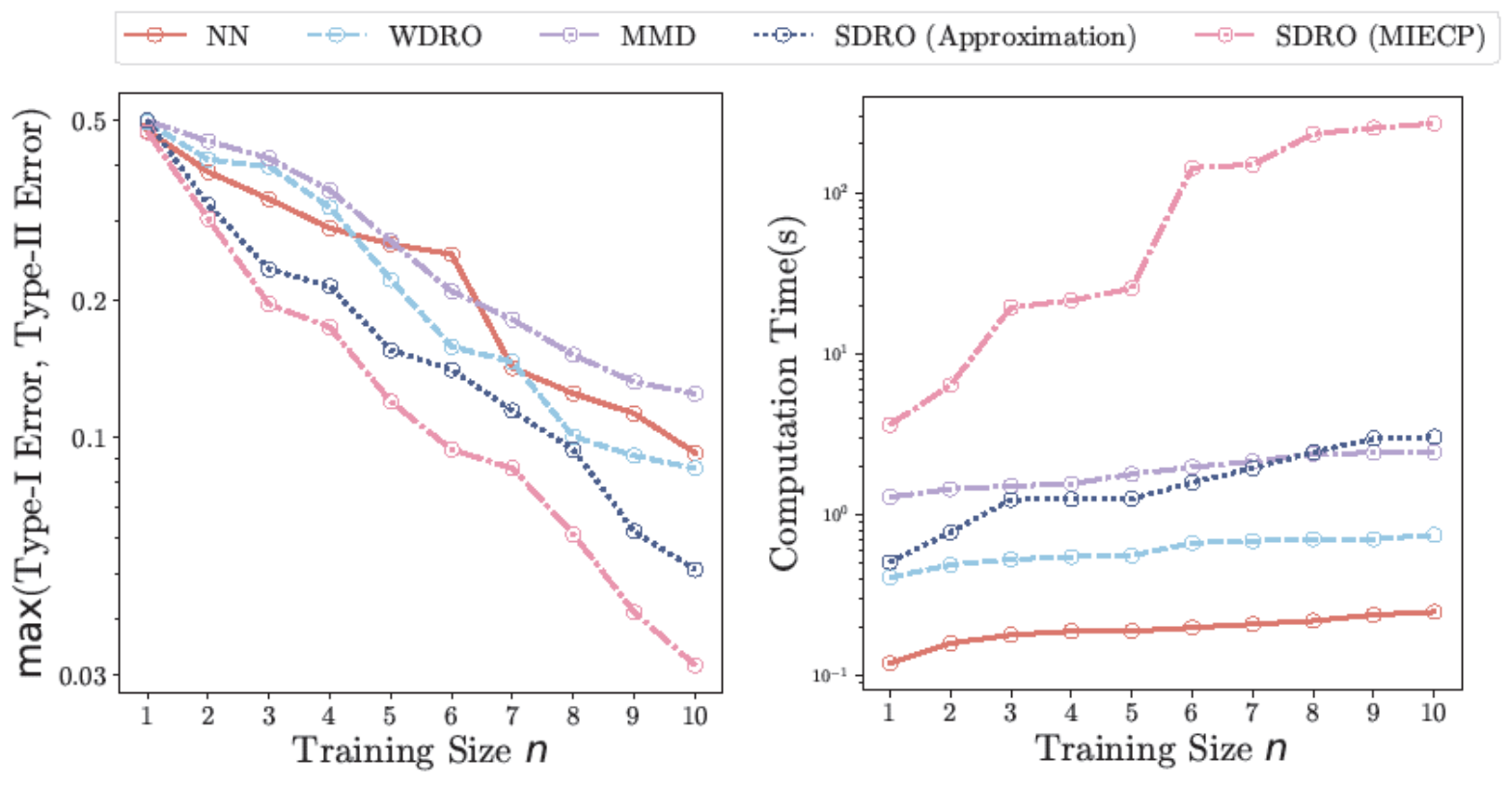}
    \caption{
    Testing performance (left plot) and computation time for various testing approaches on the HDGM dataset.}
    \label{fig:sync}
\end{figure}

Next, we perform experiments on synthetic high-dimensional Gaussian mixture~(HDGM) datasets following the similar setup as in \cite{wang2022data}, except that we use the maximum of misclassification rates on two hypotheses as the performance metric.
As illustrated in Fig.~\ref{fig:sync}, it becomes evident that our framework with either an exact or approximation algorithm attains the best performance in this example.
This achievement is noteworthy, considering that the MIECP reformulation takes longer to solve.

Finally, we apply our approximation algorithm to two commonly used image datasets: 
CIFAR-10~\cite{cifar10} and MNIST~\cite{lecun1998mnist}, and also examine our method on two medical diagnosis datasets: Lung Cancer~\cite{lung-cancer-dataset} and Sepsis~\cite{wang2022improving} (whose data are collected from Emroy and Grady hospitals in private).
We report the parameters (training and testing sample size and data dimension) of those datasets in Table~\ref{tab:para:class}.
The testing performance for various baseline methods is reported in Table~\ref{tab:acc-results}, indicating that our proposed approach outperforms other baselines in terms of the maximum type-I error and type-II error on all data sets.

\begin{table*}[!t]
\centering
\caption{Comparison of testing performance (maximum of type-I and type-II error) using real-world datasets}
\label{tab:acc-results}
\begin{tabular}{c|cccccccccccccccccccc}
\toprule[1pt]\midrule[0.4pt]
Method & 
$\begin{array}{c}
\mbox{MNIST}\\
\mbox{(Label $1$ and $2$)}
\end{array}$
& $\begin{array}{c}
\mbox{CIFAR-10}\\
\mbox{(Label $1$ and $2$)}
\end{array}$ & $\begin{array}{c}
\mbox{Lung Cancer}\\
\mbox{(Label $1$ and $2$)}
\end{array}$ & Sepsis\\
\midrule[1pt]
NN & 0.310 & 0.456 & 0.400 & 0.385\\
WDRO & 0.129 & 0.232 & \textbf{0.300} & 0.256\\
MMD & 0.356 & 0.445 & 0.500 & 0.297\\
\textbf{SDRO (Approximation)} & \textbf{0.0912} & \textbf{0.133} & \textbf{0.300} & \textbf{0.223}
\\
\midrule[0.4pt]\bottomrule[1pt]
\end{tabular}%
\end{table*}

\begin{table*}[!t]
\centering
\caption{Values of classification parameters for real-world datasets}
\label{tab:para:class}
\begin{tabular}{c|cccccccccccccccccccc}
\toprule[1pt]\midrule[0.4pt]
Parameters & MNIST~(Label $1$ and $2$) & CIFAR-10~(Label $1$ and $2$) & Lung Cancer~(Label $1$ and $2$) & Sepsis\\
\midrule[1pt]
Training Size & 50 & 50 & 12 & 20000\\
Testing Size & 2115 & 2000 & 10 & 3662\\
Data Dimension & 784 & 1024 & 56 & 39
\\
\midrule[0.4pt]\bottomrule[1pt]
\end{tabular}%
\end{table*}

\section{Concluding Remarks}
Our proposed framework opens avenues for further research exploration and refinement.
First, it is of research interest to develop more scalable optimization algorithms for solving the MIECP formulation~\eqref{Eq:MIECP} and consider enhanced convex approximations. 
One promising direction, inspired from reference~\cite{jiang2023also}, is to utilize probabilistic constraints for feasibility checks and seek the optimal detector using CVaR approximation during iterations of bisection search.
Second, there is potential for relaxing the technical assumptions when showing the regularization effects for robust hypothesis testing.
Finally, we are curious to explore the statistical guarantees for choosing the hyper-parameters of the robust testing model, including the radii and regularization parameters.

\section*{Acknowledgement}

This work is partially supported by an NSF CAREER CCF-1650913, NSF DMS-2134037, CMMI-2015787, CMMI-2112533, DMS-1938106, DMS-1830210, and the Coca-Cola Foundation.

\bibliographystyle{IEEEtran}
\bibliography{shortbib.bib}

\appendices
\section{Proofs of Technical Results}

\begin{IEEEproof}[Proof of Proposition~\ref{Pro:direct:app}]
    According to the definition of $\mathcal{F}_K$, it holds that there exists $\theta^*(\cdot)\in L^2(\pi_0)$ such that $T^*(x) = \mathbb{E}_{\omega\sim \pi_0}[\theta^*(\omega)\phi(x;\omega)]$.
Define $\bar{\theta}^* = (\theta^*(\omega_1),\ldots,\theta^*(\omega_D))$ and the approximation function $T(x)=\inp{\bar{\theta}^*}{\Phi(x)}=\frac{1}{D}\sum_{i=1}^D\theta^*(\omega_i)\phi(x;\omega_i)$ clearly belongs to $\mathcal{F}_D$.
For $i\in[D]$, define the function 
\[
X_i = \theta^*(\omega_i)\phi(\cdot;\omega_i) - T^*(\cdot).
\]
Also, define the functional $f(X_1,\ldots,X_D) = \|\frac{1}{D}\sum_iX_i\|_{L^2}$.
It can be checked that $\mathbb{E}_{\omega_i}[X_i]=0$, and the bounded difference condition is satisfied:
\[
\left| 
f(X_1,\ldots,X_D)
-
f(X_1',\ldots,X_D')
\right|\le \frac{1}{D}\|X_i - X_i'\|_{L^2}\le \frac{2M\|T\|_{\infty}}{D}.
\]
By Mcdiamard's inequality, it holds that
\[
\mathrm{Pr}\left\{ 
f(X_1,\ldots,X_D)\ge \mathbb{E}[f(X_1,\ldots,X_D)] + \delta
\right\}\le \exp\left( 
-\frac{\epsilon^2D}{2M^2\|T\|_{\infty}^2}
\right)
\]
Next, we find that 
\begin{align*}
 \mathbb{E}[f(X_1,\ldots,X_D)]&\le \Big( 
 \mathbb{E}[f^2(X_1,\ldots,X_D)]
 \Big)^{1/2}\\
 &\le \left( 
\int  \mathbb{E}_{\omega_i}\big[\frac{1}{D}\sum_i\theta^*(\omega_i)\phi(x;\omega_i) - T^*(x)\big]^2\diff x
 \right)^{1/2}\\
 &\le \frac{1}{\sqrt{D}}
 \left( 
\int  \mathbb{E}_{\omega_1}\big[\theta^*(\omega_1)\phi(x;\omega_1) - T^*(x)\big]^2\diff x
 \right)^{1/2}\\
 &=\frac{1}{\sqrt{D}}
 \left( 
\int  \Var_{\omega_1}\big[\theta^*(\omega_1)\phi(x;\omega_1) - T^*(x)\big]\diff x
 \right)^{1/2}\\ 
 &\le  \frac{1}{\sqrt{D}}
 \left( 
\int \mathbb{E}_{\omega_1}\big[\theta^*(\omega_1)\phi(x;\omega_1)\big]^2\diff x
 \right)^{1/2}
 =
 \frac{1}{\sqrt{D}} \left( 
\mathbb{E}_{\omega_1}  \|\theta^*(\omega_1)\phi(\cdot;\omega_1)\|_{L_2}^2
 \right)^{1/2}\\
 &=\frac{M\|T^*\|_{\mathcal{H}_K}}{\sqrt{D}}.
\end{align*}
Substituting the expression of $f(X_1,\ldots,X_D)$, we conclude that 
\[
\mathrm{Pr}\left\{
\|T^* - T\|_{L^2}\ge 
\frac{M}{\sqrt{D}}
\left( 
\|T^*\|_{\mathcal{H}_K}
+
\sqrt{2\|T\|_{\infty}^2\log\frac{1}{\alpha}}
\right)
\right\}\le \alpha.
\]
\end{IEEEproof}%

\begin{IEEEproof}[Proof of Proposition~\ref{Theorem:Sinkhorn}]
The proof of this proposition is a direct application of Theorem~1 in \cite{wang2021sinkhorn}.
\end{IEEEproof}

We present the following technical lemma that can be used to show Theorem~\ref{Thm:MIECP}.
\begin{lemma}\label{Lemma:conic}
For fixed $\lambda>0$, the set 
\[
\left\{
(t,z):~t\ge \lambda\log\left(
\frac{1}{m}\sum_{j\in[m]}e^{z_j/\lambda}
\right)
\right\}
\]
has exponential conic representation:
\[
(a, z, t):~\left\{
\begin{aligned}
\lambda&\ge \frac{1}{m}\sum_{j\in[m]}a_{j}\\ 
a_j&\ge \lambda \exp\left( 
\frac{z_j-t}{\lambda}
\right),\quad \forall j\in[m]
\end{aligned}
\right.
\]
\end{lemma}
\begin{IEEEproof}[Proof of Theorem~\ref{Thm:MIECP}]
Recall Problem~\eqref{Eq:CCP:testing:duality:app} is a finite-dimensional optimization:
\begin{subequations}
\begin{align}
\min_{
\substack{
\theta\in\mathbb{R}^D, \|\theta\|_2\le 1, \\
s\ge0, 
\lambda_k\ge0, k\in\{1,2\}
}}&\quad s\\
\mbox{s.t.}&\quad 
\lambda_k\orho_k + \frac{\lambda_k\orho_k}{n}\sum_{i\in[n]}
\log\left[ \frac{1}{m}\sum_{j\in[m]}
e^{\frac{\bI \{(-1)^{k+1}\inp{\theta}{\Phi(y_{i,j}^k)}<0 \}}{\lambda_k\Reg_k}}
\right]\le s,\quad k=1,2.\label{Eq:constr:16b}
\end{align}
\end{subequations}
For $k=1,2$, one can introduce slack variables $z_{i,j}^k\in\{0,1\}$ to represent the indicator function $\bI \{(-1)^{k+1}\inp{\theta}{\Phi(y_{i,j}^k)}<0 \}$, which leads to the following equivalent constraints of \eqref{Eq:constr:16b}:
\begin{subequations}
\begin{align}
\lambda_k\orho_k + \frac{\lambda_k\orho_k}{n}\sum_{i\in[n]}
\log\left[ \frac{1}{m}\sum_{j\in[m]}
e^{\frac{z_{i,j}^k}{\lambda_k\Reg_k}}
\right]&\le s\label{Eq:constr:17a}\\
(-1)^{k+1}\inp{\theta}{\Phi(y_{i,j}^k)}&\le M_{i,j}^k(1-z_{i,j}^k).\label{Eq:constr:17b}
\end{align}
\end{subequations}
Specifically, $z_{i,j}^k=1$ implies $(-1)^{k+1}\inp{\theta}{\Phi(y_{i,j}^k)}\le 0$.
On the alternative case where $z_{i,j}^k=0$, according to the definition of constant $M_{i,j}^k$, 
the constraint \eqref{Eq:constr:17b} does not have any restriction on $\theta$.

Next, we have the conic reformulation on the constraint \eqref{Eq:constr:17a}.
We introduce the slack variable $t_i^k$ to represent $\lambda_k\Reg_k\log\left[ \frac{1}{m}\sum_{j\in[m]}
e^{\frac{z_{i,j}^k}{\lambda_k\Reg_k}}
\right]$, leading to the following equivalent constraints:
\begin{subequations}
\begin{align}
\lambda_k\orho_k + \frac{1}{n}\sum_{i\in[n]}
t_i^k&\le s\label{Eq:constr:18a}\\
\lambda_k\Reg_k\log\left[ \frac{1}{m}\sum_{j\in[m]}
e^{\frac{z_{i,j}^k}{\lambda_k\Reg_k}}
\right]&\le t_i^k.\label{Eq:constr:18b}
\end{align}
Further, by Lemma~\ref{Lemma:conic}, we have the exponential conic reformulation on the constraint \eqref{Eq:constr:18b}:
\begin{align}
\lambda_k\Reg_k&\ge \frac{1}{m}\sum_{j\in[m]}a_{i,j}^k,
\quad i\in[n], \label{Eq:constr:18c}\\
(\lambda_k\Reg_k, a_{i,j}^k, z_{i,j}^k - t_i^k)&\in \mathcal{K}_{\exp},\qquad\qquad i\in[n], j\in[m]. \label{Eq:constr:18d}
\end{align}
In summary, combining constraints \eqref{Eq:constr:17b}, \eqref{Eq:constr:18a}, \eqref{Eq:constr:18c}, \eqref{Eq:constr:18d} gives the desired reformulation.
\end{subequations}

\end{IEEEproof}

\begin{IEEEproof}[Proof of Proposition~\ref{Pro:consistent}]
Since $\orho_k>0, k\in\{1,2\}$, the optimal multipliers for \eqref{Eq:CCP:testing:duality} and \eqref{Eq:CCP:testing:duality:app} are always bounded.
Assume $\lambda_k\le M$ for a constant $M>0$.
Then without loss of generality, we focus on the following \emph{compact} constraint set for Problem~\eqref{Eq:CCP:testing:duality} and \eqref{Eq:CCP:testing:duality:app}:
\[
\mathcal{M}:=\Big\{
(\theta, s, \lambda_1, \lambda_2):~
\|\theta\|_2\le 1, 0\le s\le 1, 0\le \lambda_k\le M, k=1,2
\Big\}.
\]
We first argue the SAA problem~\eqref{Eq:CCP:testing:duality:app} has non-empty set of optimal solutions for sufficiently large $m$.
Based on our technical assumption, there exists $(\theta, s, \lambda_1, \lambda_2)\in\mathcal{M}$ and $F_k(\theta, \lambda_k)<s^*$.
Due to the law of large number, $\widetilde{F}_k(\theta, \lambda_k)\to F_k(\theta, \lambda_k)$.
Consequently, the SAA problem has a feasible solution with probability $1$~(w.p.1) for large $m$.
Since $\widetilde{F}_k$ is lower semi-continuous, the feasible set of SAA problem is closed and hence compact. 
Therefore, the set of optimal solutions to the SAA problem is non-empty w.p.1 for large $m$.

For a given $\delta>0$, let $(\theta', s', \lambda_1', \lambda_2')$ be a point in $\mathcal{M}$ and is sufficiently close to $(\theta^*, s^*, \lambda_1^*, \lambda_2^*)$ such that 
\[
\widetilde{F}_k(\theta', \lambda_k')<s^*,
s'\le s^* + \delta.
\]
Consequently, w.p.1 it holds that
\[
\limsup_{m\to\infty}\widehat{\vartheta}^*\le s'\le s^*+\delta=\vartheta^*+\delta,
\]
where the first inequality is due to the feasibility of $(\theta', s', \lambda_1', \lambda_2')$ in the SAA problem.
Since $\delta>0$ is an arbitrarily small number, it follows that
\[
\limsup_{m\to\infty}\widehat{\vartheta}^*\le\vartheta^*,\quad\text{w.p.$1$}.
\]
On the other hand, let $(\theta', s', \lambda_1', \lambda_2')$ be an optimal solution to the SAA problem.
Due to the compactness of the constraint set, one can assume, by passing to a subsequence if necessary, that $(\theta', s', \lambda_1', \lambda_2')$ converges to a point $(\bar{\theta}, \bar{s}, \bar{\lambda_1}, \bar{\lambda_2})$ w.p. $1$ as $m\to\infty$.
Based on the uniform convergence theorem provided in \cite[Proposition~5.29]{shapiro2021lectures}, w.p. 1 we have the epi-graphical convergence result that $\widetilde{F}_k(\cdot,\cdot)\to F_k(\cdot,\cdot)$.
Hence, 
\[
\liminf_{m\to\infty}\widehat{\vartheta}^*\ge\vartheta^*,\quad\text{w.p.$1$}.
\]
Combining both cases gives the desired result.
\end{IEEEproof}

Before showing the proof of Proposition~\ref{Proposition:sinkhorn:reg}, we define the notion of Sinkhorn regularizer:
\begin{equation}
\mathcal{R}_{\hP}(\rho; \Reg) = \sup_{\mathbb{P}:~\cW_{\Reg}(\bP, \hP)\le \rho}~\bP(E)
-
\mathbb{E}_{x\sim\hP}[\bQ_{x,\Reg}(E)].
\end{equation}
\begin{IEEEproof}[Proof of Proposition~\ref{Proposition:sinkhorn:reg}]
Based on the duality result in Theorem~\ref{Theorem:Sinkhorn}, it holds that
\[
\mathcal{R}_{\hP}(\rho; \Reg) = \inf_{\lambda\ge0}~\Big\{ 
\lambda\orho + 
\lambda\Reg\mathbb{E}_{x\sim\hP}\left[ 
\log(1 + \bQ_{x,\Reg}(E)(e^{1/(\lambda\Reg)} - 1)) - \bQ_{x,\Reg}(E)/(\lambda\Reg)
\right]
\Big\}
\]
We take $\underline{\mathcal{R}}_{\hP}(\rho; \Reg)=(\orho/\Reg)^{-1/2}\mathcal{R}_{\hP}(\rho; \Reg)$ and then
\begin{align*}
\underline{\mathcal{R}}_{\hP}(\rho; \Reg)&=
\inf_{\lambda\ge0}~\left\{ 
\lambda\orho^{1/2}\Reg^{1/2} + \lambda\orho^{-1/2}\Reg^{3/2}\mathbb{E}_{x\sim\hP}\left[ 
\log(1 + \bQ_{x,\Reg}(E)(e^{1/(\lambda\Reg)} - 1)) - \bQ_{x,\Reg}(E)/(\lambda\Reg)
\right]
\right\}\\ 
&=\inf_{\lambda\ge0}~\left\{ 
\lambda + \frac{\lambda}{\orho/\Reg}
\mathbb{E}_{x\sim\hP}\left[ 
\log(1 + \bQ_{x,\Reg}(E)(e^{(\orho/\Reg)^{1/2}/\lambda} - 1)) - \bQ_{x,\Reg}(E)\cdot (\orho/\Reg)^{1/2}/\lambda
\right]
\right\}.
\end{align*}
We first provide the lower bound on $\underline{\mathcal{R}}_{\hP}(\rho; \Reg)$:
\begin{align*}
\underline{\mathcal{R}}_{\hP}(\rho; \Reg)&\ge 
\inf_{\lambda\ge0}~\Bigg\{ 
\lambda + \frac{\lambda}{\orho/\Reg}
\mathbb{E}_{x\sim\hP}\Big[ 
\bQ_{x,\Reg}(E)(e^{(\orho/\Reg)^{1/2}/\lambda} - 1)\\ &\qquad\qquad\qquad - \frac{1}{2}\bQ_{x,\Reg}(E)^2(e^{(\orho/\Reg)^{1/2}/\lambda} - 1)^2 - \bQ_{x,\Reg}(E)\cdot (\orho/\Reg)^{1/2}/\lambda
\Big]
\Bigg\}\\
&=\inf_{\lambda\ge0}~\Bigg\{ 
\lambda + \frac{\lambda}{\orho/\Reg}
\mathbb{E}_{x\sim\hP}\Big[ 
\bQ_{x,\Reg}(E)(e^{(\orho/\Reg)^{1/2}/\lambda} - 1 - (\orho/\Reg)^{1/2}/\lambda
)\\ &\qquad\qquad\qquad - \frac{1}{2}\bQ_{x,\Reg}(E)^2(e^{(\orho/\Reg)^{1/2}/\lambda} - 1)^2
\Big]
\Bigg\}\\
&\ge\inf_{\lambda\ge0}~\Bigg\{ 
\lambda + \frac{\lambda}{\orho/\Reg}
\mathbb{E}_{x\sim\hP}\Big[ 
\frac{\orho/\Reg}{2\lambda^2}\bQ_{x,\Reg}(E)
-
\frac{\orho/\Reg}{2\lambda^2}\bQ_{x,\Reg}(E)^2\Big]
\Bigg\} = 2^{1/2}\sigma(E;\hP,\Reg),
\end{align*}
where the first inequality is based on the relation $\log(1+x)\ge x - \frac{x^2}{2}$, and the second inequality is based on the relation $e^x-1\ge x + \frac{x^2}{2}\ge x$.
Next, we provide the upper bound on $\underline{\mathcal{R}}_{\hP}(\rho; \Reg)$.
We will plug in the multiplier
\[
\lambda_0 = \argmin_{\lambda\ge0}~\left\{ 
\lambda + \frac{\lambda}{\orho/\Reg}
\mathbb{E}_{x\sim\hP}\Big[ 
\frac{\orho/\Reg}{2\lambda^2}\bQ_{x,\Reg}(E)
-
\frac{\orho/\Reg}{2\lambda^2}\bQ_{x,\Reg}(E)^2\Big]
\right\}=\frac{\sigma(E;\hP,\Reg)}{\sqrt{2}}>0.
\]
We define $\delta:=\orho/\Reg$ and re-write
\[
\underline{\mathcal{R}}_{\hP}(\rho; \Reg) = \inf_{\lambda\ge0}~\left\{ 
\lambda + \frac{\lambda}{\delta}
\mathbb{E}_{x\sim\hP}\left[ 
\log(1 + \bQ_{x,\Reg}(E)(e^{\delta^{1/2}/\lambda} - 1)) - \bQ_{x,\Reg}(E)\cdot \delta^{1/2}/\lambda
\right]
\right\}.
\]
Since  $\lambda_0>0$, the family of continuous functions
\[
s_{\delta}(x) = \frac{\lambda_0}{\delta}
\left[ 
\log(1 + \bQ_{x,\Reg}(E)(e^{\delta^{1/2}/\lambda_0} - 1)) - \bQ_{x,\Reg}(E)\cdot \delta^{1/2}/\lambda_0
\right]
\]
converges uniformly on compact sets to
\[
s_0(x) = \frac{1}{2\lambda_0}\sigma^2(E;\hP,\Reg),
\]
Therefore, we obtain that
\begin{align*}
\underline{\mathcal{R}}_{\hP}(\rho; \Reg) &= \inf_{\lambda\ge0}~\left\{ 
\lambda + \frac{\lambda}{\delta}
\mathbb{E}_{x\sim\hP}\left[ 
\log(1 + \bQ_{x,\Reg}(E)(e^{\delta^{1/2}/\lambda} - 1)) - \bQ_{x,\Reg}(E)\cdot \delta^{1/2}/\lambda
\right]
\right\}\\ 
&\le \lambda_0 + \mathbb{E}_{x\sim\hP}[s_{\delta}(x)]\to \lambda_0 + \mathbb{E}_{x\sim\hP}[s_{0}(x)]=2^{1/2}\sigma(E;\hP,\Reg).
\end{align*}
The upper bound holds uniformly over probability measures $\hP$ that satisfies $\inf_{\Reg>0}\sigma^2(E;\hP,\Reg)\ge b_0$.
Combining upper and lower bounds gives the desired result.
\end{IEEEproof}

\begin{IEEEproof}[Proof of Proposition~\ref{Pro:cW:W}]
It is equivalent to show 
\[
\inf_{\mathbb{P}:~\cW_{\Reg}(\bP, \hP)\le \rho}~\bP(E)
=
\inf_{\mathbb{P}:~\cW_{0}(\bP, \hP)\le \rho}~\bP(E) + O(\Reg/\orho),
\]
since it always holds that
\[
\sup_{\mathbb{P}:~\cW_{\Reg}(\bP, \hP)\le \rho}~\bP(E)
=
1-\inf_{\mathbb{P}:~\cW_{\Reg}(\bP, \hP)\le \rho}~\mathbb{P}(E^c),\quad\forall\Reg\ge0.
\]
According to the definition, 
\[
\inf_{\mathbb{P}:~\cW_{\Reg}(\bP, \hP)\le \rho}~\bP(E) = \sup_{\lambda\ge0}~\left\{ 
-\lambda\orho +\mathbb{E}_{x\sim\hP}\left[ 
-\lambda\Reg\log\Big( 
\bQ_{x,\Reg}(E^c)(1 - e^{-1/(\lambda\Reg)}) + e^{-1/(\lambda\Reg)}
\Big)
\right]
\right\},
\]
where $\bQ_{x,\Reg}(E^c) = \Big[\int 1_{E^c}(z)e^{-\|z-x\|_2^2/2\Reg}\diff z\Big]/\Big[\int e^{-\|z-x\|_2^2/2\Reg}\diff z\Big]$.
\begin{itemize}
    \item 
We first show that the optimal multiplier satisfies $\lambda\Reg = O(\Reg/\orho)$ under the scaling regime that $\Reg/\orho\to0$.
When using the change of variable $a=\lambda\Reg$, the dual objective above becomes
\[
-a\cdot \frac{\orho}{\Reg} + \mathbb{E}_{x\sim\hP}\left[ 
-a\log\Big( 
\bQ_{x,\Reg}(E^c)(1 - e^{-1/a}) + e^{-1/a}
\Big)
\right].
\]
Suppose on the contrary that $\frac{a}{\Reg/\orho}\to\infty$, then the dual objective is lower unbounded, which contradicts the optimality of $\lambda$.
   \item
   According to Laplace's method~\cite{erdelyi1956asymptotic}, as $\Reg\to0$, it holds that
\[
\bQ_{x,\Reg}(E^c)=\left\{
\begin{aligned}
1+o(1),&\quad\text{if }x\in E^c,\\
(1+o(1))\cdot\exp\left( 
-\frac{d_{E^c}^2(x)}{2\Reg}
\right),&\quad\text{if }x\in E.
\end{aligned}
\right.
\]
As a consequence, 
\begin{multline*}
\inf_{\mathbb{P}:~\cW_{\Reg}(\bP, \hP)\le \rho}~\bP(E) = \sup_{\lambda\ge0}~\Bigg\{ 
-\lambda\orho +
\\
\mathbb{E}_{x\sim\hP}\left[ 
-1_E(x)\cdot \lambda\Reg\log\Bigg( 
\exp\left( 
-\frac{d_{E^c}^2(x)}{2\Reg}
\right)\cdot (1 - e^{-1/(\lambda\Reg)}) + e^{-1/(\lambda\Reg)}
\Bigg)
\right] + o(\lambda\Reg)
\Bigg\}.
\end{multline*}
It is worth mentioning that 
\begin{align*}
&\log\Bigg( 
\exp\left( 
-\frac{d_{E^c}^2(x)}{2\Reg}
\right)\cdot (1 - e^{-1/(\lambda\Reg)}) + e^{-1/(\lambda\Reg)}
\Bigg)\\
=&\log\Bigg( 
\exp\left( 
-\frac{d_{E^c}^2(x)}{2\Reg}
\right) + e^{-1/(\lambda\Reg)}
\Bigg) + \log\left( 
1 + \frac{\exp\left( 
-\frac{d_{E^c}^2(x)}{2\Reg}
\right)}{1 + e^{1/(\lambda\Reg)}\exp\left( 
-\frac{d_{E^c}^2(x)}{2\Reg}
\right)}
\right)\\
=&\log\Bigg( 
\exp\left( 
-\frac{d_{E^c}^2(x)}{2\Reg}
\right) + e^{-1/(\lambda\Reg)}
\Bigg) + O\left(\exp\left( 
-\frac{d_{E^c}^2(x)}{2\Reg}
\right)\right)\\
=&\frac{1}{\lambda\Reg}\max\left(-1, -\frac{\lambda d_{E^c}^2(x)}{2}\right) + O\left(\exp\left( 
-\frac{d_{E^c}^2(x)}{2\Reg}
\right)\right) + O(1),
\end{align*}
where the second equality is because $\log(1 + A/(1 + B))\le A/(1+B)\le A$ for sufficiently small constants $A$ and $B$, and the last equality is based on the fact that 
\[
\frac{1}{\lambda\Reg}\max\{a,b\}
\le 
\log(e^{a/(\lambda\Reg)} + e^{b/(\lambda\Reg)})\le \frac{1}{\lambda\Reg}\max\{a,b\} + \log2.
\]
Consequently,
\[
\inf_{\mathbb{P}:~\cW_{\Reg}(\bP, \hP)\le \rho}~\bP(E) = \sup_{\lambda\ge0}~\left\{ 
-\lambda\orho +\mathbb{E}_{x\sim\hP}\left[ 
1_E(x)\cdot \min\left(1, \frac{\lambda d_{E^c}^2(x)}{2}\right)
\right] + O(\lambda\Reg)
\right\}.
\]
It should be noted that $\lambda\Reg = O(\Reg/\orho)$, and therefore, the convergence result holds.
The proof is completed by checking from \cite[Theorem~2]{yangwasserstein} that
\[
\sup_{\lambda\ge0}~\left\{ 
-\lambda\orho +\mathbb{E}_{x\sim\hP}\left[ 
1_E(x)\cdot \min\left(1, \frac{\lambda d_{E^c}^2(x)}{2}\right)
\right]
\right\}=\inf_{\mathbb{P}:~\cW_{0}(\bP, \hP)\le \rho}~\bP(E).
\]
\end{itemize}
\end{IEEEproof}

\clearpage
\section{Implementation Details}\label{Appendix:B-A}
In this section, we provide details for implementing our numerical study.
\begin{itemize}
    \item 
We adopt the linear program formulation from \cite{xie2021robust} to perform the robust hypothesis testing with Wasserstein uncertainty sets, called the \textbf{WDRO} approach.
   \item
   We adopt the finite-dimensional optimization formulation from \cite[Theorem~1]{sun2023kernel} to solve the robust hypothesis testing with MMD uncertainty sets, denoted as the \textbf{MMD} approach.
   \item
We train a detector using neural networks following the literature~\cite{cheng2020classification}, called the \textbf{NN} approach.
    \item
When considering the convex approximation formulation~\eqref{Eq:convex:relaxation} for Sinkhorn robust hypothesis testing~(called \textbf{SDRO~(Approximation)}), the key is to solve the optimization subproblem~\eqref{Eq:opt:feasibility:sub}, which can further be reformulated as
\begin{equation}\label{Eq:sim:G}
\min_{\substack{\theta\in\mathbb{R}^D, \|\theta\|_2\le 1, \beta_k\le 0, \\
\lambda_k\ge0, k\in\{1,2\},\\
\tau\in\Delta_2\subseteq \mathbb{R}^2_+
}}\left\{
\tau_1G_1(s,\beta_1,\lambda_1) + \tau_2G_2(s,\beta_2,\lambda_2)
\right\},
\end{equation}
where the probability simplex $\Delta_2=\{\tau\in  \mathbb{R}^2_+:~\tau_1+\tau_2=1\}$.
We use the following stochastic approximation-based algorithm to find a near-optimal solution for this convex optimization problem. 
The objective defined in \eqref{Eq:obj:T} serves as a biased estimator of the objective defined in \eqref{Eq:sim:G}, whereas the bias vanishes as $m\to\infty$.
We will provide the theoretical analysis regarding this biased stochastic optimization algorithm in future work.
\begin{algorithm}[H]
\caption{
Stochastic Approximation for Solving Problem~\eqref{Eq:sim:G} 
} 
\label{Alg:BSGD}
\begin{algorithmic}[1]%
\REQUIRE{Maximum Iteration $T$, batch size $m$, probability level $s$}
\FOR{$t=1,\ldots,T-1$}
\STATE{Sample $x_k\sim\hP_k$ and $\{y_{j}^k\}_{j=1}^m\sim \bQ_{x,\Reg_k}$ for $k=1,2$.}
\STATE{Compute the gradient of the objective
\begin{multline}\label{Eq:obj:T}
\widehat{T}_s(\theta, \beta_1, \beta_2, \lambda_1, \lambda_2, \tau_1,\tau_2)
\\=
\sum_{k=1}^2\tau_k\left\{s\beta_k + %
\Bigg[
\lambda_k\orho_k + 
\lambda_k\Reg_k\log
\left( 
\frac{1}{m}\sum_{j=1}^m
e^{[(-1)^{k}\inp{\theta}{\Phi(y_j^k)} - \beta_k]_+/(\lambda_k\Reg_k)}
\right)
\Bigg]\right\}.
\end{multline}
}
\STATE{Update $(\theta, \beta_1, \beta_2, \lambda_1, \lambda_2, \tau_1,\tau_2)$ using projected gradient descent.}
\ENDFOR\\
\textbf{Return} $(\theta, \beta_1, \beta_2, \lambda_1, \lambda_2, \tau_1,\tau_2)$.
\end{algorithmic}
\end{algorithm}

\end{itemize}

\end{document}